\pdfoutput=1

\documentclass[11pt]{article}

\usepackage[]{emnlp2021}

\usepackage{times}
\usepackage{latexsym}

\usepackage[T1]{fontenc}

\usepackage[utf8]{inputenc}

\usepackage{microtype}

\usepackage[toc,page]{appendix}
\usepackage{colortbl}

\hypersetup{
	linkcolor=magenta,
	filecolor=magenta,
}

\usepackage{times}
\usepackage{latexsym}
\usepackage{graphicx}
\usepackage{float}
\usepackage{amsmath}
\usepackage{amsfonts}
\usepackage{subfigure}
\usepackage{algorithmic}
\usepackage[ruled,vlined]{algorithm2e}

\usepackage{booktabs}
\usepackage{placeins}
\usepackage{makecell}
\usepackage{bm}
\usepackage{multirow}
\usepackage{multicol}
\usepackage{arydshln}
\usepackage{url}
\usepackage{tablefootnote}
\usepackage[normalem]{ulem}

\usepackage{CJKutf8} 

\usepackage{amssymb}
\usepackage{pifont}
\newcommand{\cmark}{\ding{51}}%
\newcommand{\xmark}{\ding{55}}%

\newcommand{\scaleboxrate}{0.885}

\definecolor{mycolor1}{HTML}{495464}
\definecolor{mycolor2}{HTML}{bbbfca}
\definecolor{mycolor3}{HTML}{e8e8e8}
\definecolor{mycolor4}{HTML}{f4f4f2}
\newcommand{\cclq}[1]{\cellcolor{mycolor1!60}{#1}}
\newcommand{\cclw}[1]{\cellcolor{mycolor2}{#1}}
\newcommand{\ccle}[1]{\cellcolor{mycolor4}{#1}}
\newcommand{\cclr}[1]{\cellcolor{mycolor3}{#1}}

\newcommand{\modelref}[2]{#1 \cite{#2}}

\newcommand{\figref}[1]{Figure~\ref{#1}}
\newcommand{\tabref}[1]{Table~\ref{#1}}

\newcommand{\myparagraph}[1]{\vspace{-3pt}\paragraph{#1}}

\newcommand{\ie}{i.e., }

\newcommand{\mycaption}[1]{\vspace{-9pt}\caption{#1}\vspace{-9pt}}

\renewcommand\footnotemark{}
\title{{D}y{L}ex: Incorporating Dynamic Lexicons into BERT for Sequence Labeling}

\author{
	Baojun Wang$^{1*}\thanks{*Equal contribution.}$, {\bf Zhao Zhang$^{2*}$,} {\bf Kun Xu$^{2*}$,} {\bf Guang-Yuan Hao$^{3}$,} {\bf Yuyang Zhang$^{1}$} \\  {\bf Lifeng Shang$^{1}$,} {\bf Linlin Li$^{2}$,} {\bf Xiao Chen$^{1}$,} {\bf Xin Jiang$^{1}$,} {\bf Qun Liu$^{1}$} \\
	\ \ $^1$Huawei Noah's Ark Lab 
	$^{2}$Huawei Technologies Co., Ltd. \\
	$^3$The Hong Kong University of Science and Technology\\
	\{puking.w,zhangzhao54,xukun24,zhangyuyang4\}@huawei.com \\
	\{Shang.Lifeng,lynn.lilinlin,chen.xiao2,Jiang.Xin,qun.liu\}@huawei.com \\
	guangyuanhao@outlook.com \\
}

\begin{document}
	
	\maketitle
	\begin{abstract}
	Incorporating lexical knowledge into deep learning models has been proved to be very effective for sequence labeling tasks. However, previous works commonly have difficulty dealing with large-scale dynamic lexicons which often cause excessive matching noise and problems of frequent updates. 
    In this paper, we propose DyLex, a plug-in lexicon incorporation approach for BERT based sequence labeling tasks. Instead of leveraging embeddings of words in the lexicon as in conventional methods, we adopt \emph{word-agnostic tag embeddings} to avoid re-training the representation while updating the lexicon. Moreover, we employ an effective supervised lexical knowledge denoising method to smooth out matching noise. Finally, we introduce a \emph{col-wise attention} based knowledge fusion mechanism to guarantee the pluggability of the proposed framework. Experiments on ten datasets of three tasks show that the proposed framework achieves new SOTA, even with very large scale lexicons\footnote{\url{https://github.com/huawei-noah/noah-research/tree/master/NLP/dylex}}.
\end{abstract}
	
	\section{Introduction}
    Sequence labeling is the task of assigning categorical labels to a text sequence. Many conventional NLP tasks, such as named entity recognition~(NER), Chinese word segmentation~(CWS), and slot-filling based natural language understanding~(NLU), can be formalized as the sequence labeling problem. The deep learning methods, especially the recently proposed BERT and its variants, have achieved great success in such sequence labeling tasks. However, the BERT-based methods are generally built based on word-piece or character embeddings. The word information~(e.g., word {\it boundary} or {\it type}) is not fully exploited, which makes it difficult to accurately determine the entity boundary or correctly predict entity type. 
    
     \begin{figure}
        \centering
        \scalebox{0.59}{
            \includegraphics{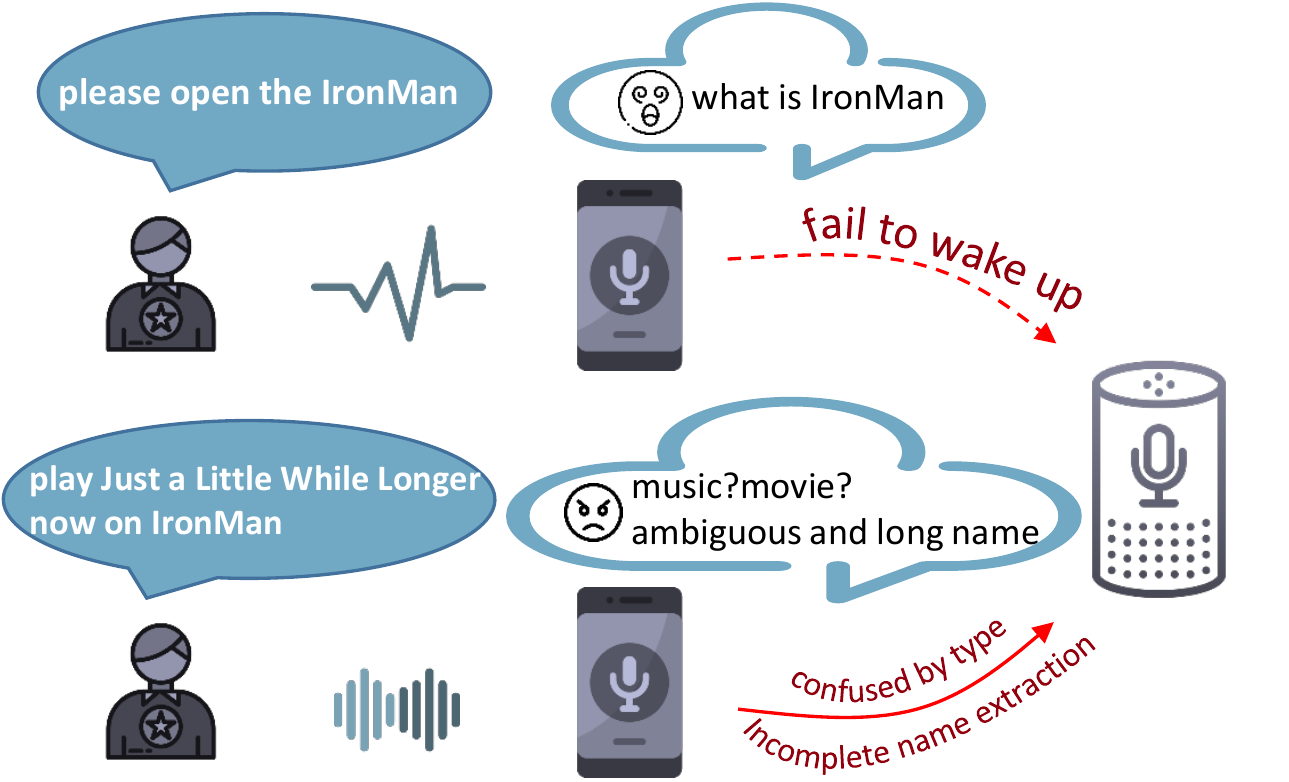}
        }
        \mycaption{
            Iron Man can be a name of a smart device or a movie and the system would be unable to react properly upon ``Please play Iron Man'' from a user. Another case as ``Play just a little while longer now on Iron Man'' requires the system to classify ``Play'' between music and movie domains, and whether ``now'' should be combined with ``just a little while longer'' as a whole.
        }
        \label{fig:shili}
    \end{figure}
    
    As shown in Figure \ref{fig:shili}, it is infeasible to understand user's utterance correctly without using deterministic domain knowledge that ``{\it Iron Man}'' is the alias of a Smart Speaker or ``{\it just a little while longer}'' is a famous song. In commercial systems, the lexicon is widely used as an effective way to store various domain knowledge. In practice, the size of a lexicon can range from ten to a few million, and we usually need to update the contents of lexicons frequently, which dramatically increases the difficulty of incorporating lexicons into deep models. In this work, we will study how to {\it effectively incorporate large-scale dynamic lexicons} into BERT-based sequence labeling models. 
    
    Recent works on incorporating lexicon knowledge~
    \cite{zhang2018chinese,ding2019neural,xiaofeng2020,li2020flat} can be summarized as follows.
    First, they match an input sentence with several lexicons to obtain all matched items. 
    Second, leveraging the matched item information through modifying the structure of the transformer layer or the feature representation layer.
    However, 1) current methods normally learn additional embeddings of the words in the lexicons, which bring us a challenge - if the lexicons get updated, the model must be re-trained; 
    2) they only use the words in the lexicon but ignore the category of words, which is important for many tasks.
    
    In this paper, we propose a general framework DyLex for incorporating frequently updated lexicons into sequence labeling models. 
    The matching results of the input are reconstructed as a word-agnostic tag sequence.
    Then we design a supervised knowledge denoising module to smooth out noisy matches, and the remaining matches are further used as additional feature input for knowledge fusion. This step is based on a {\it col-wise attention} to seamlessly fuse word-piece embeddings of input sentence and the lexicon features. Moreover, since we do not explicitly learn embeddings of the words in lexicons, there is no need to retrain the entire model when updating the lexicons.

    We conduct extensive experiments with the CWS, NER, and NLU tasks on various datasets. The results show that our model consistently outperforms the strong baselines and achieves new state-of-the-art results.

    We summarize the contribution of this work as follows:
    
    \begin{itemize}
     \item [1)] 
     We propose a general framework for effectively introducing external lexical knowledge into sequence labeling tasks. Our framework supports dynamic updates of lexicons to facilitate industrial deployment.
     \item [2)]
     We devise a novel knowledge denoising module to make full use of large-scale lexicons.
     \item [3)]
     Our framework outperforms strong baselines and achieves SOTA results on three different sequence labeling tasks.
    \end{itemize}
	
\section{Approach}
	
\begin{figure*}[h]
		\centering
		\includegraphics[width=1.0\textwidth]{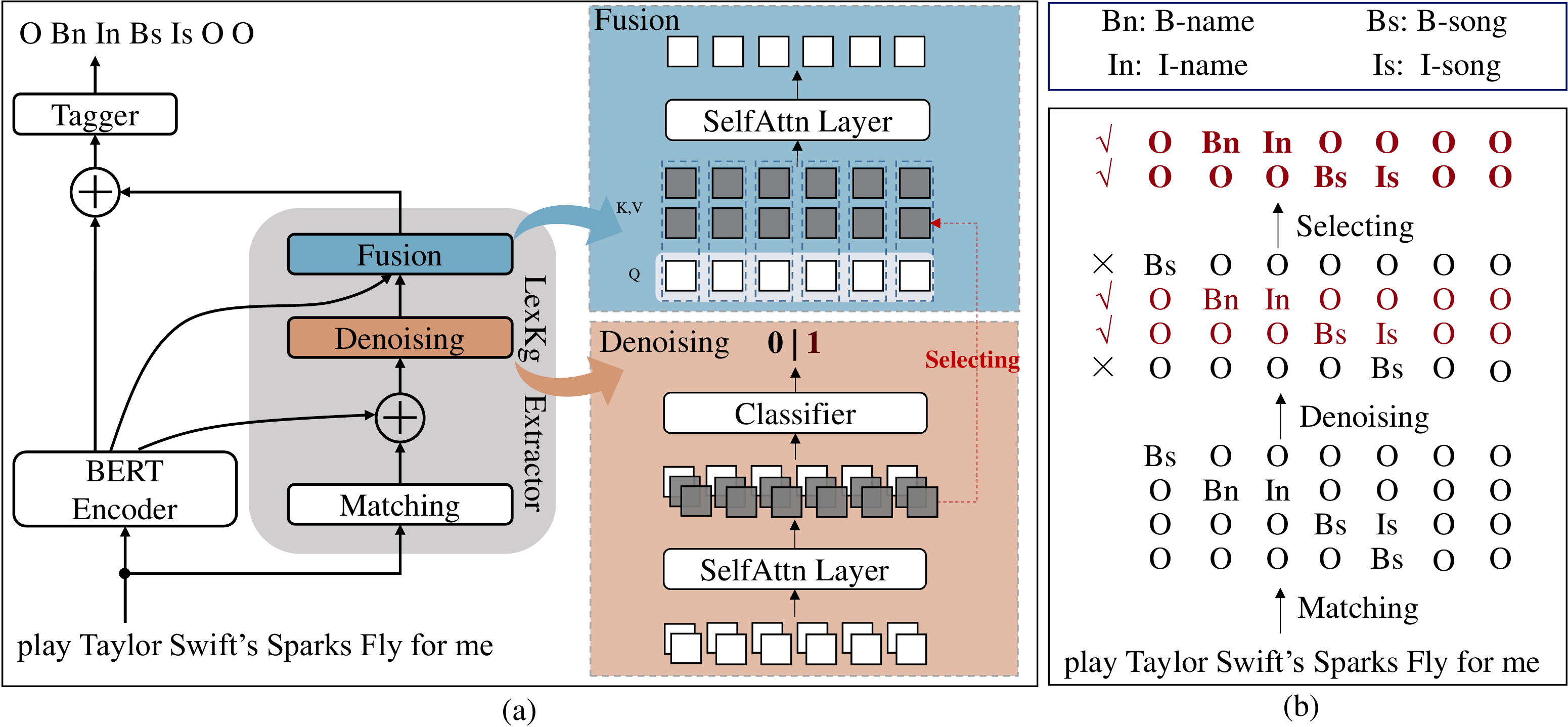}
		\mycaption{
		    (a) The overall architecture of the proposed DyLex framework, it consists of two parts, namely BERT-based sequence tagger and LexKg Extractor. The Extractor has three submodules: the Matching, the Denoising and the Fusing. (b) A concrete example of lexicon matching and denoising.
		}
		\label{fig:overview-of-model}
	\end{figure*}

	In this section, we will present how to incorporate large-scale lexicons into BERT. As illustrated in \figref{fig:overview-of-model}, the proposed DyLex framework contains two parts, namely the {\it BERT-based sequence tagger} and {\it Lexicon Knowledge extractor}. The Lexicon Knowledge~(LexKg) extractor has three submodules: Matching, Denoising and Fusing.
	
\subsection{BERT as Encoder}
	\citet{google-bert} introduces a new language representation model called BERT, which has become the building blocks of modern NLP systems. BERT is constructed based on transformer \cite{vaswani2017attention} layer, which employs multi-head attention to perform self-attention over a sequence individually and finally applies concatenation and linear transformation to the results from each head. Every single head attention in multi-head attention is calculated in a scaled dot product form:
	
	{
	\small
	\begin{equation}
	    \mathrm{Att}(Q, K, V) =\mathrm{softmax}\left(\frac{QK^{\mathrm{T}}}{\sqrt{d_{k}}}\right)V,
	\end{equation}
	}where $Q, K, V$ are input matrices, respectively. Then self-attention can be formalized as:
	
	{
	\small
	\begin{equation}
	    \mathrm{SelfAtt}(X) =\mathrm{Att}(XW_Q, XW_K, XW_V),
	\end{equation}
	}where $W_Q, W_K, W_V$ are parameter matrices to be learned.
	
\subsection{The LexKg Extractor}

\begin{algorithm}[t]
	\begin{normalsize}	
		\SetAlgoNoLine  
		\caption{Fast Matching} 
		\label{appendix:fast-matching}
		\KwIn{Trie Tree $Tr$ built from Lexicon $D$, utterance $U$} 
		\KwOut{Candidate tag sequence $T$} 
		$T$ = []\;
		\For{$i=0;i \le length(U)$}
		{
			\For{$j=i;j \le length(U)$} 
			{ 
				\uIf{$U[i:j]$ in $T$}
				{
					\tcp{reconstruct tags}
					$tags \leftarrow$ get\_tags($i,j,U,Tr$) \\
					$tags$ append to $T$\;
				}
			} 
		}
		return $T$\; 
		\BlankLine
		\SetKwFunction{FC}{get\_tags}
		\SetKwProg{Fn}{Function}{:}{}
		\Fn{\FC{$i$, $j$, u, Tr}}{
			\tcp{get lexical class}
			$class \leftarrow Tr$.match($u[i:j]$) \\
			$tags$ $\leftarrow$ label $u[i:j]$ with $class$, other position char label $O$  \\
			\KwRet tags \\
		}
	\end{normalsize}
\end{algorithm}

\myparagraph{Matching} 
    Conventional methods normally learn additional word embeddings of lexicons to incorporate lexicon knowledge, thus it is required to retrain the entire model once the lexicons are updated.
    Our method is independent of the lexicon size and lexicon word content by designing a word-agnostic representation.  
	Specifically, the Matching module takes a word sequence as input, then uses a prefix tree-based fast matching algorithm (see algorithm~\ref{appendix:fast-matching}) to quickly retrieve the lexicons, and finally produces multiple \emph{word-agnostic tag} sequences. \figref{fig:overview-of-model}~(b) shows a concrete example.

	To be detailed, we use the prefix Trie tree \cite{brass2008advanced} to store and retrieve the lexicons. 
	The non-leaf nodes of Trie are made up of word-pieces of lexicon words tokenized by BERT tokenizer,  while the leaf nodes are made up by the types of the lexicon words, namely tag name~(e.g. `B-song' and `I-song' as shown in \figref{fig:overview-of-model}~(b)).
	For each subsequence of the input, the Trie may match several different candidates. Every single match can be categorized by a tag attached with a leaf node, the rest of the sequence will be filled with `O' tags. 
	
	Formally, we denote the input sequence as ${U}$. A tag sequence obtained by fast matching is ${T^{(i)}}$, and superscript $i$ represents the index of the tag sequence. The {\it Matching} submodule can be formalized as: 
	
	\vspace{-0.5cm}
	{
		\small
		\begin{align}
		    &{E}_{u} = \mathrm{BERT}({U}) \\
			&{E}_{t}^{(i)} = \mathrm{Embedding}({T^{(i)}})\\
			&{E}_{d}^{(i)} = {E}_{t}^{(i)}+{E}_{u},
		\end{align}
	}where ${E}_{u} \in \mathbb{R}^{l \times hz}$ (here $l$ is sequence length and $hz$ is hidden size) is the representation produced by BERT encoder, ${E}_{t}^{(i)} \in \mathbb{R}^{l \times hz}$ represents the embedding of $i$-th tag sequence, and ${E}_{d}^{(i)} \in \mathbb{R}^{l \times hz}$ is the corresponding output of this module.

	\myparagraph{Denoising}
	The proposed fast matching algorithm can quickly obtain all potential matched sub-sequences with the lexicons. However, due to the large scale size of the lexicon, even for an input sequence with only a few words, there may be dozens of incorrect matches. 
	Using \figref{fig:overview-of-model}~(b) as an example, only Row 2 (i.e. the matching to singer {\it Taylor Swift's}) and Row 3 (i.e. the matching to song {\it Sparks Fly}) are expected matchings, whereas all the other tag sequences contain incorrect matchings, namely the \emph{\textbf{matching nosie}} mentioned in this work, which will inevitably decrease final performance. Thus we devise a novel supervised knowledge denoising module to smooth them out.
	
	The supervising signal can be automatically derived from the golden sequence labels of the training dataset. In the example of \figref{fig:overview-of-model}~(b), each row corresponds to a single matching tag sequences, and Row  2 and Row 3 are used as positive training samples whereas negative for the other two.
	Note that, our method can still work even if the category of lexicon~(e.g. name or song) is not provided, in that case, a tag sequence degenerates to mark out a lexicon word boundary.
	
	Formally, we first get the representation of $i$-th tag sequence from its embedding ${E}_{d}^{(i)}$ with self-attention
	
	{
		\small
		\begin{align}
		{R}_{d}^{(i)} &= \mathrm{SelfAtt}({E}_{d}^{(i)}).
		\end{align}
	}When classifying each tag sequence, we also need to consider relationships among them.
	For example, Row 3 and Row 4 in \figref{fig:overview-of-model}~(b) can not be $\mathrm{True}$ at the same time since they share some contradicting spans.
	Taking that into consideration, we first concatenate the \texttt{[cls]} in ${R}_{d}$ (i.e. first column) of all tag sequences to form a matrix ${R_{cls}\in{\mathbb{R}^{{nd}\times hz}}}$, where $nd$ denotes the number of tag sequences~(e.g. its value is 4 in the example of \figref{fig:overview-of-model}~(b)) and $hz$ denotes the hidden representation size. Then we pass the matrix $R_{cls}$ to a self attention layer to model the interrelation among them, 
	
	{\small
	\begin{align}
		&{Y} = \mathrm{SelfAttn}({R_{cls}})\\
		&\mathrm{P}({Z}=\mathrm{True} \mid {R_{d}}) \ =\ \sigma (\mathrm{Linear}({Y})),
	\end{align}
	}where $\sigma$ represents the sigmoid function, and ${Z\in{\mathbb{R}^{nd}}}$ is the classification result. 
	The representation of a positively classified tag sequence is denoted as $R_d^{(i)+}$. These selected positive representations will be fused with the original BERT embedding $E_u$.
	
\myparagraph{Knowledge Fusing}	
	In this stage, our framework aims to produce a lexical knowledge enhanced representation ${E}_{k}$ by fusing BERT-based encoding ${E}_{u}$ with several selected tag sequences $R_d^+$ via the proposed {\it col-wise attention}. 
	Use the $j$-th token of input sequence as an example, we take its BERT-based representation $E_u^{(j)}$ to act as Query, and its corresponding tag representation $R_d^{(i,j)+}$ as Key and Value, then col-wise attention can formulate:
	
	{
	    \small
	    \begin{align}
	        &K=V = [R_d^{(1,j)+}; \dots; R_d^{(m,j)+}] \\
	        &E_k^{(i)} =~\mathrm{Att}(E_u^{(j)}, ~K, ~V),
	    \end{align}
	}where $m=|{R_d^+}|$. Then concatenate $E_k^{(i)}$ for all $l$ positions to get $E_k$.
	
\subsection{The Tagger}
	At last ${E}_{O}$ is produced by combining ${E}_{u}$ with ${E}_{k}$,
	and here we use a linear classification layer, as used by BERT tagger. 
	
	\vspace{-0.7cm}
	\begin{align}
	    \small
		{E}_{O} &= ~{E}_{u}+{E}_{k} \\
		{O} &= \mathrm{\sigma(Linear}( {E_{O}} ))
	\end{align}
	where $O$ is the classification result for each token.

	We can see that the proposed framework is not an intrusive method but rather pluggable. As we take the encoder's output as input and return a knowledge enhanced text representation, the original model structure is not modified.

\section{Experiments}

	We conduct experiments on several NLP tasks, including CWS~(Chinese word segmentation), NER~(named entity recognition), and NLU~(natural language understanding). 
	The experimental hyperparameter settings are listed in appendix \ref{app:hyper-param}.
	
\subsection{Primary Baselines}
	\myparagraph{BERT-based Sequence Tagger} 
		The framework uses BERT as an encoder to represent the input sequence. As can be seen in \figref{fig:overview-of-model}, we can get this baseline by removing the LexKg extractor part of DyLex.
		
	\myparagraph{Glyce \cite{meng2019glyce}}
		Glyce is the glyph-vectors for Chinese character representations. With the lexicon, it has achieved the best performance on Chinese word segmentation so far.	
		
	\myparagraph{FLAT and HSCRF+Softdict \cite{li2020flat,liu2019towards}}
		Named entity recognition can benefit greatly from lexicons. FLAT utilizes lexicons with the Lattice structure for Chinese entity recognition, and HSCRF with softdict is used for English named entity recognition, both of them have achieved strong results.

\subsection{Lexicon Construction}

    \label{lexicon-construct}
	The lexicon mentioned in this article refers to a collection, the entry of which contains item and Category. The item corresponds to the words, and the category corresponds to the type of the words. The category of words is customized according to the task. For example, the category in the NER task can be the song name. Tag is a BIO format that marks the type of a word.
	\tabref{tb:construct-dictionary} shows notation and appendix \ref{app:fragment of lexicon} is a detailed lexicon fragment.
	
\begin{table}[]
	\centering
	\scalebox{0.66}{
		\begin{tabular}{cccl}
			\toprule
			Task & Item & Category & Tag \\ 
			\midrule
			CWS & words &  -  & \makecell[l]{B: Begining of a word \\ I: Continuation of a word}  \\ 
			\midrule
			NER & words & Song name  & \makecell[l]{B-song: Begining of a song name \\ I-song: Continuation of a song name}  \\ 
			\midrule
			NLU & words & Location name  & \makecell[l]{B-loc: Begining of a location name \\ I-loc: Continuation of a location name}  \\ 
			\bottomrule
	\end{tabular}}
	\mycaption{
		Examples of lexicon's content in different tasks.
	}
	\label{tb:construct-dictionary}
\end{table}
\begin{table*}[h]
	\centering
	\setlength{\tabcolsep}{3.5mm}{
	\scalebox{\scaleboxrate}{
		\begin{tabular}{lcccccc}
			\toprule
			Methods          								&LEX	  & Weibo & MSRA  & Resume & Ontonotes & AVG  \\
			\midrule
			\modelref{BiLSTM-CRF}{huang2015bidirectional}   &\xmark   & 56.75 & 91.87 & 94.41  & 71.81 & 78.71    \\
			\modelref{TENER}{yan2019tener}            		&\xmark	  & 58.39 & 93.01 & 95.25  & 72.82 & 79.86    \\
			\modelref{BERT}{google-bert}   				    &\xmark	  & 68.20 & 94.95 & 95.53  & 80.14 & 84.70    \\
			\midrule
			\modelref{LSTM+ExSoftWord}{peng2019simplify}  	&\cmark	  & 56.02 & 92.38 & 95.43  & 72.40 & 79.05    \\
			\modelref{Lattice-LSTM}{zhang2018chinese} 	    &\cmark	  & 58.79 & 93.18 & 94.46  & 73.88 & 80.07    \\
			\modelref{LR-CNN}{gui2019cnn}           		&\cmark	  & 59.92 & 93.71 & 95.11  & 74.45 & 80.79    \\
			\modelref{FLAT+BERT+CRF}{li2020flat}     &\cmark  & 68.55 & 96.09 & 95.86  & \textbf{81.82} & 85.58   \\
			\midrule
			DyLex        &\cmark	& \textbf{71.12} & \textbf{96.49} & \textbf{95.99}  & 81.48 &\textbf{86.27}   \\
			\bottomrule
		\end{tabular}
	}}
	\caption{F1 scores of different methods on Chinese NER dataset. AVG stands for the average of each row.}
	\label{tb:ner-cn-result}
\end{table*}
\begin{table*}[]
	\centering
	\setlength{\tabcolsep}{4.2mm}{
	\scalebox{\scaleboxrate}{
		\begin{tabular}{p{7.0cm}p{1.4cm}<{\centering}p{2cm}<{\centering}p{2cm}<{\centering}c}
			\toprule
			Methods                      				&LEX  & Conll2003   & OntoNotes5.0 & AVG     \\
			\midrule
			\modelref{BiLSTM-CRF}{huang2015bidirectional} &\xmark  & 91.03     & 86.28      & 88.65  \\
			\modelref{TENER}{yan2019tener}              &\xmark    & 91.33     & 88.43      & 89.88  \\
			\modelref{LSTM-CNNs}{chiu2016named}        	&\xmark	   & 91.62     & 86.28      & 88.95  \\
			\modelref{BERT}{google-bert}                &\xmark    & 92.40     & 89.13      & 90.76  \\
			\modelref{CSE}{akbik2018contextual}		    &\xmark	   & 92.72	   & 89.71      & 91.40  \\
			\midrule
			\modelref{SENNA}{collobert2011natural} 		&\cmark	  & 89.56     & -          & -       \\
			\modelref{JERL}{luo2015joint}         		&\cmark	  & 91.20     & -          & -       \\
			\modelref{ID-CNN}{strubell2017fast}         &\cmark   & 90.54     & 86.84      & 88.69   \\
			\modelref{GRN}{chen2019grn}                 &\cmark   & 91.44     & 87.67      & 89.55   \\
			\modelref{HSCRF}{liu2019towards}			&\cmark	  & 92.75     & 89.94      & 91.34   \\
			\modelref{LUKE}{yamada2020luke}			    &\cmark	  & 94.30     & -          & -       \\
			\midrule
			DyLex                   &\cmark	  & \textbf{94.30} & \textbf{90.19} & \textbf{92.25}     \\
			\bottomrule
		\end{tabular}
	}
	}
	\mycaption{ F1 scores of different methods on English NER dataset. The setting is the same with Table\ref{tb:ner-cn-result}.Note that LUKE incorporate the entity information during the pre-training phase.}
	\label{tb:ner-en-result}
\end{table*}
	
	The lexicon tag mentioned above is used to mark word categories, namely the value in the lexicon, which is strongly related to the task. 
	Figure~\ref{fig:overview-of-model}(b) and the `Tag' column in Table~\ref{tb:construct-dictionary} display some examples.
	
	The lexicons used in our experiments are consistent with the ones used in baseline methods. In the NLU task, since there has not been any related work with using lexicons, we extract labeled spans from the training corpus and merge them with the lexicon used in NER task. The lexicon sizes used in our experiments are listed in appendix \ref{appendix:lexicon_size_exp}.

\subsection{Task1: Chinese Word Segmentation}
    
\begin{table}[] 
	\centering
	\scalebox{0.88}{
		\begin{tabular}{lccc}
			\toprule
			Model & LEX & PKU & CITYU \\
			\midrule
			\citet{yang2017neural} 			&\xmark		& 96.30			& 96.94 \\
			\citet{ma2018state}				&\xmark		& 96.10			& 97.23 \\
			\citet{huang2019toward} 		&\xmark		& 96.60			& 97.60 \\
			\modelref{BERT}{google-bert} 	&\xmark		& 96.50 		& 97.60 \\
			\modelref{Glyce}{meng2019glyce} &\cmark		& 96.70 		& 97.90 \\
			
			\midrule
			DyLex							&\cmark 	& \textbf{97.14}& \textbf{98.60} \\
			\bottomrule
		\end{tabular}
	}
	\mycaption{
		F1 Score on PKU and CITYU datasets.
	}	
	\label{tb:cws-result}
\end{table}

	CWS aims to divide a sentence into meaningful chunks. It is a primary task for Chinese text processing. 
	Using lexicons in CWS tasks is a commonly used operation. Brand new words and internet buzzwords emerge every day, and it is essential to add these words into lexicons for better performance.
	
	In this work, we experiment on two popular CWS datasets, i.e., PKU and CITYU\cite{emerson}. The lexicon used in this experiment is consistent with jieba word segmentation lexicon\footnote{\url{https://github.com/fxsjy/jieba}}, which consists of a simplified Chinese lexicon from jieba and an extra traditional Chinese lexicon from Taiwan version of jieba.
	We converted all traditional Chinese into simplified Chinese for all lexicons and datasets.
	
	To fairly compare our model with the SOTA models, we use the same settings on dataset split with \citet{meng2019glyce}. 
	
\begin{table*}[htbp]
	\centering
	\setlength{\tabcolsep}{3mm}{
	\scalebox{\scaleboxrate}{
		\begin{tabular}{p{3.1cm}ccccccccc}
			\toprule
			\multirow{2}{*}{MODELS} & \multicolumn{2}{c}{TEST} & \multicolumn{2}{c}{SINGLE} & \multicolumn{2}{c}{MULTI} & \multicolumn{2}{c}{MEDIA} & \multirow{2}{*}{DISAMB} \\
			\cmidrule(r){2-3}\cmidrule(r){4-5}\cmidrule(r){6-7}\cmidrule(r){8-9}
			& intent & slot & intent  & slot  & intent    & slot  & intent    & slot   &    \\
			\midrule
			BERT   & 96.67    & 95.12  & 13.83  & 54.66  & 77.13  & 81.22   & 95.46   & 92.88   & - \\
			DyLex  & \textbf{97.43}    & \textbf{96.65}  & \textbf{77.81}  & \textbf{92.10}  & \textbf{90.89}  & \textbf{93.03}   & \textbf{95.96}   & \textbf{95.09}   & \textbf{97.74} \\
			\bottomrule
		\end{tabular}
	}
	\caption{
		Performance on the industrial dataset (F1). 
		The TEST set is divided into three parts, SINGLE, MULTI, and MEDIA. 
		The slot in SINGLE can only correspond to one tag in lexicon, and the one in MULTI can correspond to multiple tag. The sentence in MEDIA has obvious indicator words, such as words like ``play music''.
	}
		\label{tb:nlu-industrial-rlt}
	}

\end{table*}

\begin{table*}[htbp]
	\setlength{\tabcolsep}{2.6mm}{
	\begin{center}
		\scalebox{\scaleboxrate}{
			\begin{tabular}{lccccccccc}
				\toprule
				\multirow{2}{*}{Models} &\multirow{2}{*}{LEX} & \multicolumn{3}{c}{Snips} & \multicolumn{3}{c}{ATIS} & \multirow{2}{*}{AVG}  \\
				\cmidrule(r){3-5}\cmidrule(r){6-8}
				& & Intent & Slot & match$_{sen}$ & Intent & Slot & match$_{sen}$ \\
				\midrule
				Atten-joint~\cite{liu2016attention}&\xmark & 96.7 & 87.8 & 74.1 & 91.1 & 94.2 & 78.9 & 87.13 \\
				Slot-Gated~\cite{goo2018slot} &\xmark		& 97.0 & 88.8 & 75.5 & 94.1 & 95.2 & 82.6 & 88.86 \\
				SF-ID~\cite{niu2019novel}&\xmark 			& 97.4 & 92.2 & 80.5 & 97.7 & 95.8 & 86.7 & 91.71 \\
				Joint BERT~\cite{chen2019bert}&\xmark 	& 98.6 & 97.0 & 92.8 & 97.5 & $\bm{96.1}$ & 88.2 & 95.03 \\
				HSCRF$^*$~\cite{liu2019towards} &\cmark   & 98.7 & 97.6 & 93.1 & 97.7 & 96.0 & 88.4 & 95.25 \\
				\midrule
				DyLex &\cmark 	& $\bm{99.8}$ & $\bm{99.1}$ & $\bm{98.1}$ & $\bm{98.2}$ & ${95.7}$ & $\bm{88.5}$ & $\bm{96.52}$ \\
				\bottomrule
			\end{tabular}
		}
	\end{center}
	\mycaption{NLU performance on Snips and ATIS datasets. The metrics are intent classification accuracy, slot filling F1, and sentence-level semantic frame accuracy (\%). 
	The results marked with * are reported from our recurrence.
	}
	\label{tb:pub-NLU-rlt}
	}
\end{table*}

	As shown in \tabref{tb:cws-result}, our method outperforms all the other compared baselines.
	Compared with Glyce, which is a strong baseline, our method obtains improvement of 0.44\% and 0.7\% on PKU and CITYU respectively.

\subsection{Task2: Named Entity Recognition}
		
	Named entity recognition is a typical sequence labeling task, and it heavily relies on external knowledge. 
	Incorporating lexicon as external knowledge can help determine the span and type of entities.
	To fully verify the capability of the proposed framework in NER, 
	we evaluate our framework on 
	Ontonotes~\cite{trove.nla.gov.au/work/192067053}, 
	MSRA~\cite{levow2006third}, 
	Resume~\cite{zhang2018chinese}, 
	and Weibo~\cite{peng2015named,he2016f} 
	for Chinese, and 
	Conll2003~\cite{sang2003introduction} 
	and Ontonotes~\cite{pradhan2017ontonotes} 
	for English. 
	The statistics of these datasets are detailed in \tabref{tb:overview-ner-dataset}. 
	The lexicon used in Chinese NER tasks is the same as \citet{li2020flat}, and the one in English is the same as \citet{liu2019towards}. 

	We first evaluate our framework on the Chinese datasets, and the results are shown in \tabref{tb:ner-cn-result}. 
	Except for the Ontonotes, our approach achieves the best results over all methods with  lexicons, averagely 0.69\% higher than FLAT.
	Compared with BERT, which is the best method without using lexicon, our approach improves even more dramatically, with 1.57\% higher.
	
	We evaluate our framework on two English datasets (\ie{Conll2003, OnotNotes5.0}). The conclusion is similar to Chinese Datasets, as shown in \tabref{tb:ner-en-result}. Comparing with the HSCRF and CSE, our method is 0.91\% and 0.85\% higher on average, with and without lexicon respectively.
	\modelref{LUKE}{yamada2020luke} scores the same as our method on the conll 2003 data set, and also uses information related to entities. they achieved it through pre-training, which is orthogonal to our method.
	
		\begin{figure*}[h]
    	\centering
    	\includegraphics[width=1.0\textwidth]{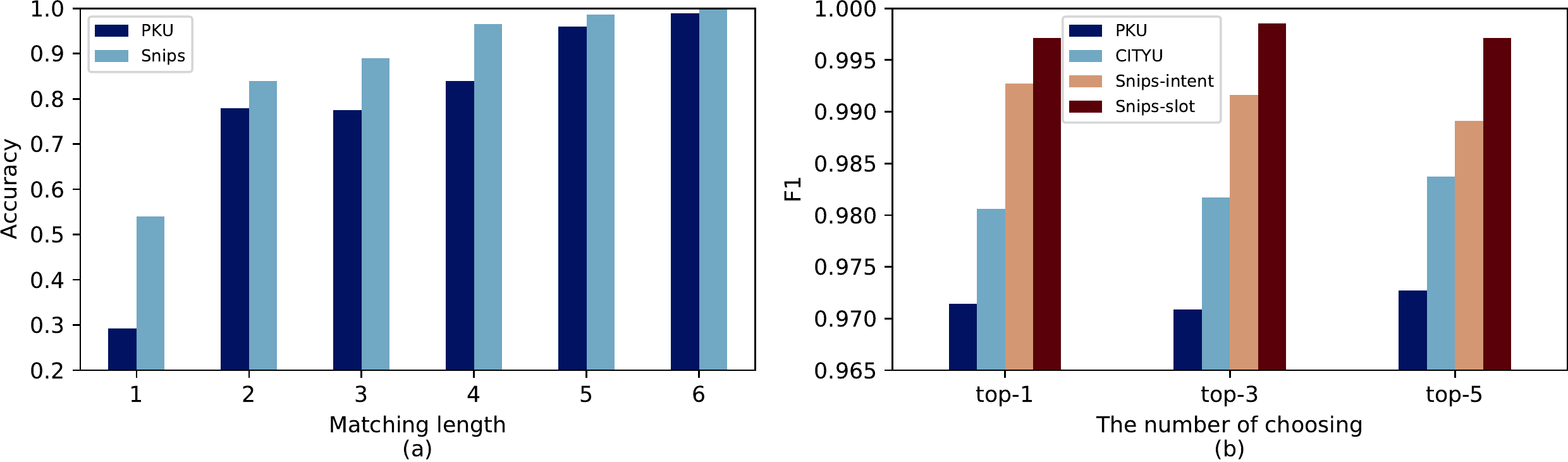}
    	\caption{
    		(a) The Influence of matching length (the x-axis represents the matched word's length, and the y-axis represents the proportion of correct results in all matching results). (b) F1 score on top-n candidates by reverse order of match length. (top-n means fetching n the longest matching results).
    	}
    	\label{fig:match-length-and-num}
    \end{figure*}

\subsection{Task3: Natural Language Understanding}

	NLU is a more challenging sequence labeling task, which aims to recognize the intent of spoken language and extract slots. 
	As shown in \figref{fig:shili}, in many practical application scenarios, one cannot tell the real intent unless the entity is provided as prior knowledge.
	
	We evaluate the framework on an industrial data set and two public data sets. 
	The chinese industrial data set is a commercial dataset for mobile phone assistant.
	The public datasets are Snips\footnote{https://github.com/snipsco/nlubenchmark/} and ATIS \cite{tur2010left}. 
	The details of the three datasets are shown in appendix \ref{appedix:nlu-dataset}.
	
	The overall performance of our framework on the industrial dataset is listed in \tabref{tb:nlu-industrial-rlt}. For the test set, there are 0.76\% and 1.53\% improvements in intent detection and slot filling, respectively. 
	Specifically, the gain is more obvious in the SINGLE and MULTI set. 
	The BERT can not distinguish intent between ``play music'' and ``play video'' since the model lacks the prior knowledge of whether ''Love Story'' is a song or a movie. 
	In the MEDIA set, all sentences contain demonstrative words, such as ``play music [xxx]'' and ``play video [xxx]''. This type of sentence does not depend on the type of xxx. It is easy to make judgments through the demonstrative words~(i.e., music and video), but there is still a 0.5\% increase in intent detection, and the increment in the slot filling is even more obvious, reaching 2.21\%.
	
	The experimental results on Snips and ATIS are shown in Table \ref{tb:pub-NLU-rlt}, the setting follows previous works~\nocite{Huang2020NLU}~\cite{niu2019novel,goo2018slot}.
	It can be seen that our framework outperforms the other methods in all three metrics (except slot of ATIS): slot filling (F1), intent detection (Acc), and sentence accuracy (Acc), with 1.27\% higher on average than the previous best method. 
	For ATIS, the improvement is not as much as other methods. This is mainly because the dataset is relatively small and the slot is sparse, lexicons are underutilized.


\begin{table*}[]
	\centering
	\setlength{\tabcolsep}{3.2mm}{
	\scalebox{\scaleboxrate}{
	\begin{tabular}{p{2.4cm}p{2.0cm}p{2.0cm}<{\centering}p{1.5cm}<{\centering}p{1.9cm}<{\centering}p{1.5cm}<{\centering}p{1.9cm}<{\centering}}
		\toprule
		\multirow{2}{*}{Task}   & \multirow{2}{*}{Datasets} & \multirow{2}{*}{BERT} & \multicolumn{2}{c}{Exp-Dict} & \multicolumn{2}{c}{Sp-Dict} \\ \cmidrule(r){4-5}  \cmidrule(r){6-7} 
		&                           &                         & Denoising    & Dylex               & Denoising    & Dylex     \\
		\midrule 
		\multirow{2}{*}{CWS}        & PKU        & 96.50      & 97.90        & 97.14(+0.64)        & 99.26        & 98.11(+1.61)    \\
									& CITYU      & 97.60      & 97.91        & 98.06(+0.46)        & 99.14        & 98.72(+1.12)    \\ 
		\midrule
		\multirow{4}{*}{NER-Chinese} & Ontonotes & 80.14      & 97.83        & 81.48(+1.34)        & 98.37        & 82.31(+2.17)    \\ 
									& MSRA       & 94.65      & 98.10        & 96.40(+1.75)        & 98.74        & 96.85(+2.20)    \\ 
									& Resume     & 95.53      & 97.92        & 95.99(+0.46)        & 98.82        & 96.40(+0.87)    \\ 
									& Weibo      & 68.20      & 96.93        & 71.12(+2.92)        & 97.83        & 71.53(+3.33)    \\ 
		\midrule
		\multirow{2}{*}{NER-English}& Conll2003  & 92.40      & 98.66        & 94.30(+1.90)        & 98.81        & 94.44(+2.04)    \\ 
									& Ontonotes5.0 & 89.13    & 97.24        & 90.19(+1.06)        & 98.19        & 91.40(+2.27)    \\ 
		\bottomrule
	\end{tabular}}
	\caption{
		Column BERT represents the F1 on each task, the Denoising column represents the accuracy of the denoising module, and the Dylex column is the F1 of our method and its increment versus BERT. Exp-dict is the lexicon corresponding to each experiment above, and Sp-Dict indicates specialized domain-related lexicons.
	}
	\vspace{-8pt}
	\label{tb:denoising}
	}
\end{table*}
	
\section{Discussion}
\subsection{The Study of Match Length}
	\label{sec:match-length}

	Given an utterance, the FM(algorithm~\ref{appendix:fast-matching}) often produces numerous matching results for each position. On the one hand, we are not sure which result is correct. To retain the correct result, we should keep as many results as possible. On the other hand, most matching results are invalid, bringing a lot of matching noise and increasing computation cost. We have to make a balance between them. 
	As shown in the \figref{fig:match-length-and-num}(a), the longer the length is, the higher the accuracy is. 
	Based on this observation, we should select matching results by reverse order of match length.
	
	We also studied the number of selected results for each position in the sentence.
	It is more likely to keep the right matches with a larger number, but it brings more noise.
	From \figref{fig:match-length-and-num}(b), F1 on the three data sets do not increase as the number grows. Taking efficiency into account, we generally select $n=1$ or $n=2$.
	
	    \begin{figure}[h]
		\centering
		\includegraphics[width=7.4cm]{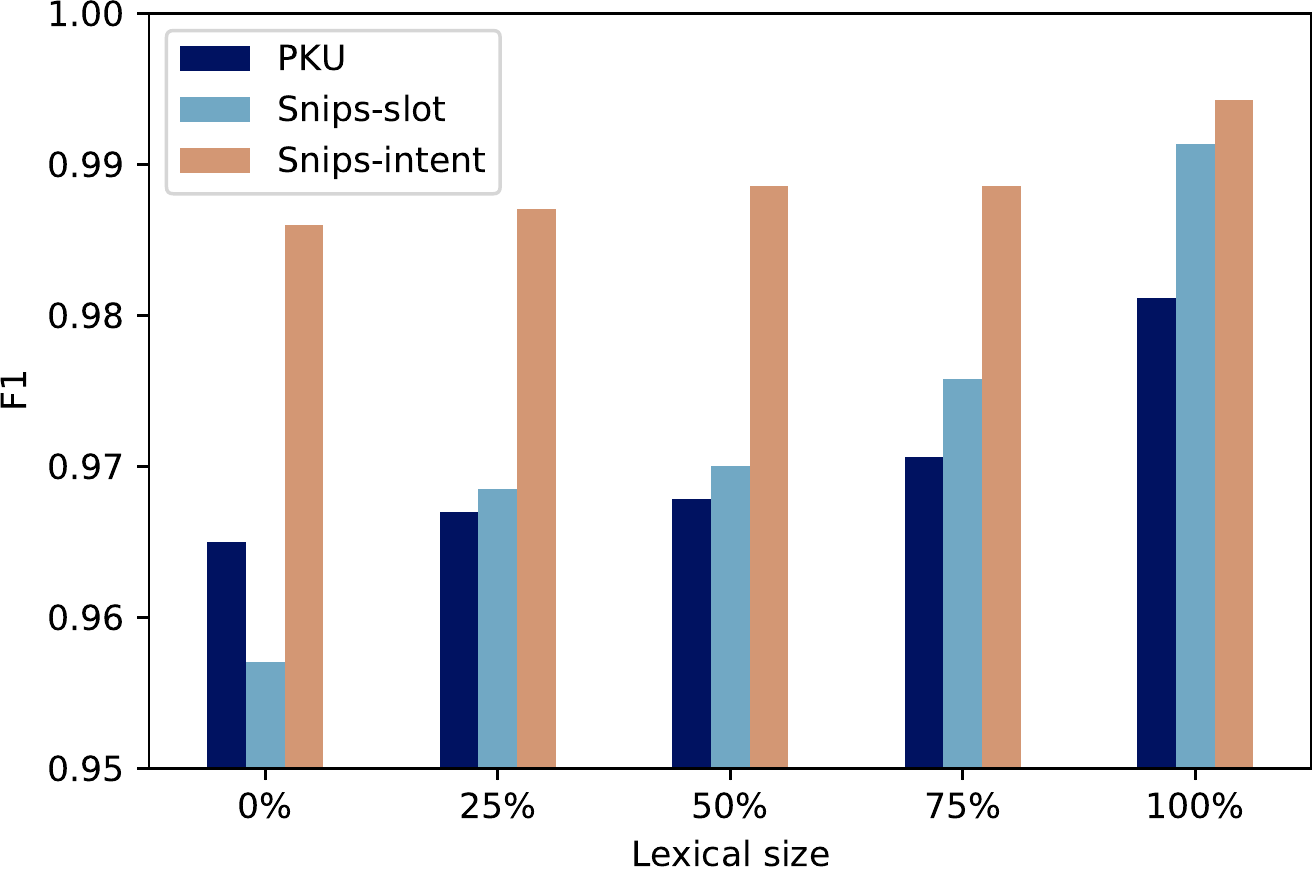}
		\mycaption{
			The F1 of different task with different lexical size. When the size is 100\%, it means using the entire lexicon in the corresponding experiment above.}
		\label{fig:on-the-fly}
	\end{figure}

	\subsection{Effect of Dynamic Lexicon}
	One advantage of our proposed method is the ability to load lexicons dynamically. Instead of using the embedding of updated lexicon entries, we only use the lexicon words' category tags.
	Thus we can expand the scale of lexicons arbitrarily without retraining. We studied the effect of lexicon size on performance. From  \figref{fig:on-the-fly}, we can see that without using a lexicon, the performance are close to the results of BERT base. With the increasing size of lexicon, the performance will also be improved.
	
	\subsection{Look Back on Denoising}
    Indistinguishable lexicon matching will bring huge noise.
	The quality of denoising will affect the performance of the model. From \tabref{tb:denoising}, we can see that whether it is Exp-Dict or Sp-Dict, the more precise the denoising, the more improvement will be achieved compared to BERT without using a lexicon. 
	The Sp-Dict here is a specialized collection of domain lexicons. For example, the lexicon only contains entities of the relevant category in the NER task, and the scale is relatively small. In this case, the matching noise brought by Sp-Dict is much smaller. From the \tabref{tb:denoising}, we can observe that the accuracy of denoising in Sp-dict is better than that in Exp-dict, which directly leads to impressive improvement in the experiment. This also confirms the importance of denoising.

	
	\subsection{Fusion in Hard or Soft Way}
	After Denoising, the results $R_{d}$ should be fused with $E_u$ for downstream tasks. The fused methods can be soft or hard. In the soft setting, all of the $R_{d}$ are weighted summed before fusing with $E_u$. The advantage of this is we can use gradient back-propagation to train the model. Different from the soft method, $E_u$ in the hard method is selected according to the threshold.
    As shown in \tabref{tb:hard-or-soft}, the overall performance of hard fusion is better since it mainly fuses more accurate results. Besides, we also adopt Teacher Forcing \cite{williams} in soft/hard methods, but it does not yield promising accuracies.
    
    \begin{table}[htbp]
	\centering
	\setlength{\tabcolsep}{3.2mm}{
	\scalebox{0.9}{
		\begin{tabular}{cp{1.0cm}<{\centering}p{1.0cm}<{\centering}p{1.0cm}<{\centering}p{1.0cm}<{\centering}}
			\toprule
			\multirow{2}{*}{Methods} & \multicolumn{2}{c}{MSRA} & \multicolumn{2}{c}{Resume} \\
			\cmidrule(r){2-3}  \cmidrule(r){4-5}
					& Exp-Dict        & Sp-Dict         & Exp-Dict        & Exp-Dict               \\
			\midrule
			Soft    & 95.26           & 95.51           & 95.13           & 94.91                \\
			Hard    & \textbf{96.40}  & \textbf{96.85}  & \textbf{95.99}  & \textbf{96.40}       \\
			\bottomrule
		\end{tabular}
	}}
	\mycaption{
		The F1 of two selecting strategies.
	}
	\label{tb:hard-or-soft}
\end{table}
	
	\section{Related Work}
	With the advance of deep learning, sequence labelling tasks, such as segmentation and NER, have achieved excellent performance. More and more methods tend to be character-based\cite{chen2006chinese,lu2016multi,dong2016character}, especially in languages, such as Chinese, Japanese, Korean, etc., that require word segmentation.  These languages do not have a natural segmentation delimiter as white space in Latin languages. Character-based input in these languages can avoid accumulation of word segmentation errors, then get better performances\cite{he2008chinese,liu2010chinese,li2014comparison}. However, the downside of the purely character-based method is that the word information is not fully exploited.
	
	To make full use of word information, incorporating a lexicon is an effective method. 
	Existing works on incorporating lexicon can be categorized as feature based, lattice based and graph based methods according to implementation complexity.
	
	\paragraph{Feature based}
	Feature based method is a simpler way. Some works directly use lexical information with simple matching features and the others use auxiliary tasks to leverage the lexical information. \citet{zhang2018neural} builds the template first and uses the template matching lexicon to build features, which help word segmentation tasks. \citet{xiaofeng2020} uses a simple lexicon matching location information as features. \citet{li2014comparison} and  \citet{peters2017semi}  adopt word-level language modeling objective and multi-task to use word information implicitly. \citet{yang2017transfer} transfer cross-domain and cross-lingual knowledge via multi-task learning. 
	
	\paragraph{Lattice based}
	Lattice based method is to use lattice structure. 
	\citet{zhang2018chinese} proposes Lattice-LSTM for incorporating word lexicons into the character-based NER model. Rather than heuristically choosing a word for the character when matching multiple words in the lexicon, they also introduce an elaborate modification to the sequence modeling layer of the LSTM-CRF model\cite{huang2015bidirectional}.
	Considering that the short path in the lattice structure will cause the word-based structure to degenerate into a character-based structure, \citet{liu2019encoding}  propose a novel word character LSTM(WC-LSTM) model to add word information via four strategies. 
	Since the lattice structure is complex and dynamic, most existing lattice-based models cannot fully utilize GPUs' parallel computation and usually have a low inference-speed. \citet{li2020flat} propose a Transformer-based model for Chinese NER, which converts the lattice structure into a flat structure.
	
	\paragraph{Graph based}
	Graph based method uses a directed graph structure to fuse lexiconal information. 
	\citet{gui2019lexicon} uses a GNN-based method to explore multiple graph-based interactions among characters, potential words, and the whole-sentence semantics to effectively alleviate the word ambiguity.
	\citet{sui2019leverage} employ a collaborative graph network to assign both self-matched and the nearest contextual lexical terms.
	To automatically learn how to incorporate multiple gazetteers into a NER system, \citet{ding2019neural} propose a novel approach based on graph neural networks with a multidigraph structure. The structure captures the information the gazetteers offer. 
	
	    \vspace{-0.1cm}
\section{Conclusion and Future Work}
	\vspace{-0.1cm}
	In this paper, we propose DyLex, a framework incorporating dynamic lexicon to improve BERT-like models' performance in sequence labeling tasks. To alleviate the problems caused by large-scale dynamic lexicons, we introduce word-agnostic tag embeddings and a knowledge denoising module. As a result, our framework outperforms the state-of-the-art works on many sequence labeling tasks.
	In future,  
	how to extend it to text classification is a challenge, since denoising corpus cannot be automatically constructed at this time.
	
	\section*{Acknowledgments}
We thank the anonymous reviewers for their insightful comments. We also appreciate the helpful discussion with the colleagues in our team.

	\bibliography{anthology,dylex}

\begin{thebibliography}{49}
\expandafter\ifx\csname natexlab\endcsname\relax\def\natexlab#1{#1}\fi

\bibitem[{Akbik et~al.(2018)Akbik, Blythe, and Vollgraf}]{akbik2018contextual}
Alan Akbik, Duncan Blythe, and Roland Vollgraf. 2018.
\newblock \href {https://aclanthology.org/C18-1139/} {Contextual string
  embeddings for sequence labeling}.
\newblock In \emph{Proceedings of the 27th International Conference on
  Computational Linguistics, {COLING} 2018, Santa Fe, New Mexico, USA, August
  20-26, 2018}, pages 1638--1649. Association for Computational Linguistics.

\bibitem[{Brass(2008)}]{brass2008advanced}
Peter Brass. 2008.
\newblock \emph{Advanced data structures}, volume 193.
\newblock Cambridge University Press Cambridge.

\bibitem[{Chen et~al.(2019{\natexlab{a}})Chen, Lin, Ding, Lou, Zhang, and
  Karlsson}]{chen2019grn}
Hui Chen, Zijia Lin, Guiguang Ding, Jianguang Lou, Yusen Zhang, and B{\"{o}}rje
  Karlsson. 2019{\natexlab{a}}.
\newblock \href {https://doi.org/10.1609/aaai.v33i01.33016236} {{GRN:} gated
  relation network to enhance convolutional neural network for named entity
  recognition}.
\newblock In \emph{The Thirty-Third {AAAI} Conference on Artificial
  Intelligence, {AAAI} 2019, The Thirty-First Innovative Applications of
  Artificial Intelligence Conference, {IAAI} 2019, The Ninth {AAAI} Symposium
  on Educational Advances in Artificial Intelligence, {EAAI} 2019, Honolulu,
  Hawaii, USA, January 27 - February 1, 2019}, pages 6236--6243. {AAAI} Press.

\bibitem[{Chen et~al.(2019{\natexlab{b}})Chen, Zhuo, and Wang}]{chen2019bert}
Qian Chen, Zhu Zhuo, and Wen Wang. 2019{\natexlab{b}}.
\newblock \href {http://arxiv.org/abs/1902.10909} {{BERT} for joint intent
  classification and slot filling}.
\newblock \emph{CoRR}, abs/1902.10909.

\bibitem[{Chen et~al.(2006)Chen, Zhang, and Isahara}]{chen2006chinese}
Wenliang Chen, Yujie Zhang, and Hitoshi Isahara. 2006.
\newblock \href {https://aclanthology.org/W06-0116/} {Chinese named entity
  recognition with conditional random fields}.
\newblock In \emph{Proceedings of the Fifth Workshop on Chinese Language
  Processing, SIGHAN@COLING/ACL 2006, Sydney, Australia, July 22-23, 2006},
  pages 118--121. Association for Computational Linguistics.

\bibitem[{Chiu and Nichols(2016)}]{chiu2016named}
Jason P.~C. Chiu and Eric Nichols. 2016.
\newblock \href {https://transacl.org/ojs/index.php/tacl/article/view/792}
  {Named entity recognition with bidirectional lstm-cnns}.
\newblock \emph{Trans. Assoc. Comput. Linguistics}, 4:357--370.

\bibitem[{Collobert et~al.(2011)Collobert, Weston, Bottou, Karlen, Kavukcuoglu,
  and Kuksa}]{collobert2011natural}
Ronan Collobert, Jason Weston, L{\'{e}}on Bottou, Michael Karlen, Koray
  Kavukcuoglu, and Pavel~P. Kuksa. 2011.
\newblock \href {http://dl.acm.org/citation.cfm?id=2078186} {Natural language
  processing (almost) from scratch}.
\newblock \emph{J. Mach. Learn. Res.}, 12:2493--2537.

\bibitem[{Devlin et~al.(2019)Devlin, Chang, Lee, and Toutanova}]{google-bert}
Jacob Devlin, Ming{-}Wei Chang, Kenton Lee, and Kristina Toutanova. 2019.
\newblock \href {https://doi.org/10.18653/v1/n19-1423} {{BERT:} pre-training of
  deep bidirectional transformers for language understanding}.
\newblock In \emph{Proceedings of the 2019 Conference of the North American
  Chapter of the Association for Computational Linguistics: Human Language
  Technologies, {NAACL-HLT} 2019, Minneapolis, MN, USA, June 2-7, 2019, Volume
  1 (Long and Short Papers)}, pages 4171--4186. Association for Computational
  Linguistics.

\bibitem[{Ding et~al.(2019)Ding, Xie, Zhang, Lu, Li, and Si}]{ding2019neural}
Ruixue Ding, Pengjun Xie, Xiaoyan Zhang, Wei Lu, Linlin Li, and Luo Si. 2019.
\newblock \href {https://doi.org/10.18653/v1/p19-1141} {A neural multi-digraph
  model for chinese {NER} with gazetteers}.
\newblock In \emph{Proceedings of the 57th Conference of the Association for
  Computational Linguistics, {ACL} 2019, Florence, Italy, July 28- August 2,
  2019, Volume 1: Long Papers}, pages 1462--1467. Association for Computational
  Linguistics.

\bibitem[{Dong et~al.(2016)Dong, Zhang, Zong, Hattori, and
  Di}]{dong2016character}
Chuanhai Dong, Jiajun Zhang, Chengqing Zong, Masanori Hattori, and Hui Di.
  2016.
\newblock \href {https://doi.org/10.1007/978-3-319-50496-4\_20}
  {Character-based {LSTM-CRF} with radical-level features for chinese named
  entity recognition}.
\newblock In \emph{Natural Language Understanding and Intelligent Applications
  - 5th {CCF} Conference on Natural Language Processing and Chinese Computing,
  {NLPCC} 2016, and 24th International Conference on Computer Processing of
  Oriental Languages, {ICCPOL} 2016, Kunming, China, December 2-6, 2016,
  Proceedings}, volume 10102 of \emph{Lecture Notes in Computer Science}, pages
  239--250. Springer.

\bibitem[{E et~al.(2019)E, Niu, Chen, and Song}]{niu2019novel}
Haihong E, Peiqing Niu, Zhongfu Chen, and Meina Song. 2019.
\newblock \href {https://doi.org/10.18653/v1/p19-1544} {A novel bi-directional
  interrelated model for joint intent detection and slot filling}.
\newblock In \emph{Proceedings of the 57th Conference of the Association for
  Computational Linguistics, {ACL} 2019, Florence, Italy, July 28- August 2,
  2019, Volume 1: Long Papers}, pages 5467--5471. Association for Computational
  Linguistics.

\bibitem[{Emerson(2005)}]{emerson}
Thomas Emerson. 2005.
\newblock \href {https://aclanthology.org/I05-3017/} {The second international
  chinese word segmentation bakeoff}.
\newblock In \emph{Proceedings of the Fourth {SIGHAN} Workshop on Chinese
  Language Processing, SIGHAN@IJCNLP 2005, Jeju Island, Korea, 14-15, 2005}.
  {ACL}.

\bibitem[{Goo et~al.(2018)Goo, Gao, Hsu, Huo, Chen, Hsu, and
  Chen}]{goo2018slot}
Chih{-}Wen Goo, Guang Gao, Yun{-}Kai Hsu, Chih{-}Li Huo, Tsung{-}Chieh Chen,
  Keng{-}Wei Hsu, and Yun{-}Nung Chen. 2018.
\newblock \href {https://doi.org/10.18653/v1/n18-2118} {Slot-gated modeling for
  joint slot filling and intent prediction}.
\newblock In \emph{Proceedings of the 2018 Conference of the North American
  Chapter of the Association for Computational Linguistics: Human Language
  Technologies, NAACL-HLT, New Orleans, Louisiana, USA, June 1-6, 2018, Volume
  2 (Short Papers)}, pages 753--757. Association for Computational Linguistics.

\bibitem[{Gui et~al.(2019{\natexlab{a}})Gui, Ma, Zhang, Zhao, Jiang, and
  Huang}]{gui2019cnn}
Tao Gui, Ruotian Ma, Qi~Zhang, Lujun Zhao, Yu{-}Gang Jiang, and Xuanjing Huang.
  2019{\natexlab{a}}.
\newblock \href {https://doi.org/10.24963/ijcai.2019/692} {Cnn-based chinese
  {NER} with lexicon rethinking}.
\newblock In \emph{Proceedings of the Twenty-Eighth International Joint
  Conference on Artificial Intelligence, {IJCAI} 2019, Macao, China, August
  10-16, 2019}, pages 4982--4988. ijcai.org.

\bibitem[{Gui et~al.(2019{\natexlab{b}})Gui, Zou, Zhang, Peng, Fu, Wei, and
  Huang}]{gui2019lexicon}
Tao Gui, Yicheng Zou, Qi~Zhang, Minlong Peng, Jinlan Fu, Zhongyu Wei, and
  Xuanjing Huang. 2019{\natexlab{b}}.
\newblock \href {https://doi.org/10.18653/v1/D19-1096} {A lexicon-based graph
  neural network for chinese {NER}}.
\newblock In \emph{Proceedings of the 2019 Conference on Empirical Methods in
  Natural Language Processing and the 9th International Joint Conference on
  Natural Language Processing, {EMNLP-IJCNLP} 2019, Hong Kong, China, November
  3-7, 2019}, pages 1040--1050. Association for Computational Linguistics.

\bibitem[{He and Sun(2017)}]{he2016f}
Hangfeng He and Xu~Sun. 2017.
\newblock \href {https://doi.org/10.18653/v1/e17-2113} {F-score driven max
  margin neural network for named entity recognition in chinese social media}.
\newblock In \emph{Proceedings of the 15th Conference of the European Chapter
  of the Association for Computational Linguistics, {EACL} 2017, Valencia,
  Spain, April 3-7, 2017, Volume 2: Short Papers}, pages 713--718. Association
  for Computational Linguistics.

\bibitem[{He and Wang(2008)}]{he2008chinese}
Jingzhou He and Houfeng Wang. 2008.
\newblock \href {https://aclanthology.org/I08-4022/} {Chinese named entity
  recognition and word segmentation based on character}.
\newblock In \emph{Third International Joint Conference on Natural Language
  Processing, {IJCNLP} 2008, Hyderabad, India, January 7-12, 2008}, pages
  128--132. The Association for Computer Linguistics.

\bibitem[{Huang et~al.(2020{\natexlab{a}})Huang, Cheng, Chen, Wang, and
  Chu}]{huang2019toward}
Weipeng Huang, Xingyi Cheng, Kunlong Chen, Taifeng Wang, and Wei Chu.
  2020{\natexlab{a}}.
\newblock \href {https://doi.org/10.18653/v1/2020.coling-main.186} {Towards
  fast and accurate neural chinese word segmentation with multi-criteria
  learning}.
\newblock In \emph{Proceedings of the 28th International Conference on
  Computational Linguistics, {COLING} 2020, Barcelona, Spain (Online), December
  8-13, 2020}, pages 2062--2072. International Committee on Computational
  Linguistics.

\bibitem[{Huang et~al.(2015)Huang, Xu, and Yu}]{huang2015bidirectional}
Zhiheng Huang, Wei Xu, and Kai Yu. 2015.
\newblock \href {http://arxiv.org/abs/1508.01991} {Bidirectional {LSTM-CRF}
  models for sequence tagging}.
\newblock \emph{CoRR}, abs/1508.01991.

\bibitem[{Huang et~al.(2020{\natexlab{b}})Huang, Liu, and Zou}]{Huang2020NLU}
Zhiqi Huang, Fenglin Liu, and Yuexian Zou. 2020{\natexlab{b}}.
\newblock \href {https://doi.org/10.18653/v1/2020.coling-main.310} {Federated
  learning for spoken language understanding}.
\newblock In \emph{Proceedings of the 28th International Conference on
  Computational Linguistics, {COLING} 2020, Barcelona, Spain (Online), December
  8-13, 2020}, pages 3467--3478. International Committee on Computational
  Linguistics.

\bibitem[{Levow(2006)}]{levow2006third}
Gina{-}Anne Levow. 2006.
\newblock \href {https://aclanthology.org/W06-0115/} {The third international
  chinese language processing bakeoff: Word segmentation and named entity
  recognition}.
\newblock In \emph{Proceedings of the Fifth Workshop on Chinese Language
  Processing, SIGHAN@COLING/ACL 2006, Sydney, Australia, July 22-23, 2006},
  pages 108--117. Association for Computational Linguistics.

\bibitem[{Li et~al.(2014)Li, Hagiwara, Li, and Ji}]{li2014comparison}
Haibo Li, Masato Hagiwara, Qi~Li, and Heng Ji. 2014.
\newblock \href
  {http://www.lrec-conf.org/proceedings/lrec2014/summaries/358.html}
  {Comparison of the impact of word segmentation on name tagging for chinese
  and japanese}.
\newblock In \emph{Proceedings of the Ninth International Conference on
  Language Resources and Evaluation, {LREC} 2014, Reykjavik, Iceland, May
  26-31, 2014}, pages 2532--2536. European Language Resources Association
  {(ELRA)}.

\bibitem[{Li et~al.(2020)Li, Yan, Qiu, and Huang}]{li2020flat}
Xiaonan Li, Hang Yan, Xipeng Qiu, and Xuanjing Huang. 2020.
\newblock \href {https://doi.org/10.18653/v1/2020.acl-main.611} {{FLAT:}
  chinese {NER} using flat-lattice transformer}.
\newblock In \emph{Proceedings of the 58th Annual Meeting of the Association
  for Computational Linguistics, {ACL} 2020, Online, July 5-10, 2020}, pages
  6836--6842. Association for Computational Linguistics.

\bibitem[{Liu and Lane(2016)}]{liu2016attention}
Bing Liu and Ian~R. Lane. 2016.
\newblock \href {https://doi.org/10.21437/Interspeech.2016-1352}
  {Attention-based recurrent neural network models for joint intent detection
  and slot filling}.
\newblock In \emph{Interspeech 2016, 17th Annual Conference of the
  International Speech Communication Association, San Francisco, CA, USA,
  September 8-12, 2016}, pages 685--689. {ISCA}.

\bibitem[{Liu et~al.(2019{\natexlab{a}})Liu, Yao, and Lin}]{liu2019towards}
Tianyu Liu, Jin{-}Ge Yao, and Chin{-}Yew Lin. 2019{\natexlab{a}}.
\newblock \href {https://doi.org/10.18653/v1/p19-1524} {Towards improving
  neural named entity recognition with gazetteers}.
\newblock In \emph{Proceedings of the 57th Conference of the Association for
  Computational Linguistics, {ACL} 2019, Florence, Italy, July 28- August 2,
  2019, Volume 1: Long Papers}, pages 5301--5307. Association for Computational
  Linguistics.

\bibitem[{Liu et~al.(2019{\natexlab{b}})Liu, Xu, Xu, Song, and
  Zu}]{liu2019encoding}
Wei Liu, Tongge Xu, QingHua Xu, Jiayu Song, and Yueran Zu. 2019{\natexlab{b}}.
\newblock \href {https://doi.org/10.18653/v1/n19-1247} {An encoding strategy
  based word-character {LSTM} for chinese {NER}}.
\newblock In \emph{Proceedings of the 2019 Conference of the North American
  Chapter of the Association for Computational Linguistics: Human Language
  Technologies, {NAACL-HLT} 2019, Minneapolis, MN, USA, June 2-7, 2019, Volume
  1 (Long and Short Papers)}, pages 2379--2389. Association for Computational
  Linguistics.

\bibitem[{Liu et~al.(2010)Liu, Zhu, and Zhao}]{liu2010chinese}
Zhangxun Liu, Conghui Zhu, and Tiejun Zhao. 2010.
\newblock \href {https://doi.org/10.1007/978-3-642-14932-0\_78} {Chinese named
  entity recognition with a sequence labeling approach: Based on characters, or
  based on words?}
\newblock In \emph{Advanced Intelligent Computing Theories and Applications.
  With Aspects of Artificial Intelligence, 6th International Conference on
  Intelligent Computing, {ICIC} 2010, Changsha, China, August 18-21, 2010.
  Proceedings}, volume 6216 of \emph{Lecture Notes in Computer Science}, pages
  634--640. Springer.

\bibitem[{Lu et~al.(2016)Lu, Zhang, and Ji}]{lu2016multi}
Yanan Lu, Yue Zhang, and Dong{-}Hong Ji. 2016.
\newblock \href
  {http://www.lrec-conf.org/proceedings/lrec2016/summaries/612.html}
  {Multi-prototype chinese character embedding}.
\newblock In \emph{Proceedings of the Tenth International Conference on
  Language Resources and Evaluation {LREC} 2016, Portoro{\v{z}}, Slovenia, May
  23-28, 2016}. European Language Resources Association {(ELRA)}.

\bibitem[{Luo et~al.(2015)Luo, Huang, Lin, and Nie}]{luo2015joint}
Gang Luo, Xiaojiang Huang, Chin{-}Yew Lin, and Zaiqing Nie. 2015.
\newblock \href {https://doi.org/10.18653/v1/d15-1104} {Joint entity
  recognition and disambiguation}.
\newblock In \emph{Proceedings of the 2015 Conference on Empirical Methods in
  Natural Language Processing, {EMNLP} 2015, Lisbon, Portugal, September 17-21,
  2015}, pages 879--888. The Association for Computational Linguistics.

\bibitem[{Ma et~al.(2018)Ma, Ganchev, and Weiss}]{ma2018state}
Ji~Ma, Kuzman Ganchev, and David Weiss. 2018.
\newblock \href {https://doi.org/10.18653/v1/d18-1529} {State-of-the-art
  chinese word segmentation with bi-lstms}.
\newblock In \emph{Proceedings of the 2018 Conference on Empirical Methods in
  Natural Language Processing, Brussels, Belgium, October 31 - November 4,
  2018}, pages 4902--4908. Association for Computational Linguistics.

\bibitem[{Ma et~al.(2020)Ma, Peng, Zhang, Wei, and Huang}]{peng2019simplify}
Ruotian Ma, Minlong Peng, Qi~Zhang, Zhongyu Wei, and Xuanjing Huang. 2020.
\newblock \href {https://doi.org/10.18653/v1/2020.acl-main.528} {Simplify the
  usage of lexicon in chinese {NER}}.
\newblock In \emph{Proceedings of the 58th Annual Meeting of the Association
  for Computational Linguistics, {ACL} 2020, Online, July 5-10, 2020}, pages
  5951--5960. Association for Computational Linguistics.

\bibitem[{Meng et~al.(2019)Meng, Wu, Wang, Li, Nie, Yin, Li, Han, Sun, and
  Li}]{meng2019glyce}
Yuxian Meng, Wei Wu, Fei Wang, Xiaoya Li, Ping Nie, Fan Yin, Muyu Li, Qinghong
  Han, Xiaofei Sun, and Jiwei Li. 2019.
\newblock Glyce: Glyph-vectors for chinese character representations.
\newblock In \emph{Advances in Neural Information Processing Systems}, pages
  2746--2757.

\bibitem[{Mu et~al.(2020)Mu, Wei, and Aiping}]{xiaofeng2020}
Xiaofeng Mu, Wang Wei, and Xu~Aiping. 2020.
\newblock Incorporating token-level dictionary feature into neural model for
  named entity recognition.
\newblock \emph{Neurocomputing}.

\bibitem[{Peng and Dredze(2015)}]{peng2015named}
Nanyun Peng and Mark Dredze. 2015.
\newblock \href {https://doi.org/10.18653/v1/d15-1064} {Named entity
  recognition for chinese social media with jointly trained embeddings}.
\newblock In \emph{Proceedings of the 2015 Conference on Empirical Methods in
  Natural Language Processing, {EMNLP} 2015, Lisbon, Portugal, September 17-21,
  2015}, pages 548--554. The Association for Computational Linguistics.

\bibitem[{Peters et~al.(2017)Peters, Ammar, Bhagavatula, and
  Power}]{peters2017semi}
Matthew~E. Peters, Waleed Ammar, Chandra Bhagavatula, and Russell Power. 2017.
\newblock \href {https://doi.org/10.18653/v1/P17-1161} {Semi-supervised
  sequence tagging with bidirectional language models}.
\newblock In \emph{Proceedings of the 55th Annual Meeting of the Association
  for Computational Linguistics, {ACL} 2017, Vancouver, Canada, July 30 -
  August 4, Volume 1: Long Papers}, pages 1756--1765. Association for
  Computational Linguistics.

\bibitem[{Pradhan and Ramshaw(2017)}]{pradhan2017ontonotes}
Sameer Pradhan and Lance Ramshaw. 2017.
\newblock Ontonotes: Large scale multi-layer, multi-lingual, distributed
  annotation.
\newblock In \emph{Handbook of Linguistic Annotation}, pages 521--554.
  Springer.

\bibitem[{Sang and Meulder(2003)}]{sang2003introduction}
Erik F. Tjong~Kim Sang and Fien~De Meulder. 2003.
\newblock \href {https://aclanthology.org/W03-0419/} {Introduction to the
  conll-2003 shared task: Language-independent named entity recognition}.
\newblock In \emph{Proceedings of the Seventh Conference on Natural Language
  Learning, CoNLL 2003, Held in cooperation with {HLT-NAACL} 2003, Edmonton,
  Canada, May 31 - June 1, 2003}, pages 142--147. {ACL}.

\bibitem[{Strubell et~al.(2017)Strubell, Verga, Belanger, and
  McCallum}]{strubell2017fast}
Emma Strubell, Patrick Verga, David Belanger, and Andrew McCallum. 2017.
\newblock \href {https://doi.org/10.18653/v1/d17-1283} {Fast and accurate
  entity recognition with iterated dilated convolutions}.
\newblock In \emph{Proceedings of the 2017 Conference on Empirical Methods in
  Natural Language Processing, {EMNLP} 2017, Copenhagen, Denmark, September
  9-11, 2017}, pages 2670--2680. Association for Computational Linguistics.

\bibitem[{Sui et~al.(2019)Sui, Chen, Liu, Zhao, and Liu}]{sui2019leverage}
Dianbo Sui, Yubo Chen, Kang Liu, Jun Zhao, and Shengping Liu. 2019.
\newblock \href {https://doi.org/10.18653/v1/D19-1396} {Leverage lexical
  knowledge for chinese named entity recognition via collaborative graph
  network}.
\newblock In \emph{Proceedings of the 2019 Conference on Empirical Methods in
  Natural Language Processing and the 9th International Joint Conference on
  Natural Language Processing, {EMNLP-IJCNLP} 2019, Hong Kong, China, November
  3-7, 2019}, pages 3828--3838. Association for Computational Linguistics.

\bibitem[{T{\"{u}}r et~al.(2010)T{\"{u}}r, Hakkani{-}T{\"{u}}r, and
  Heck}]{tur2010left}
G{\"{o}}khan T{\"{u}}r, Dilek Hakkani{-}T{\"{u}}r, and Larry~P. Heck. 2010.
\newblock \href {https://doi.org/10.1109/SLT.2010.5700816} {What is left to be
  understood in atis?}
\newblock In \emph{2010 {IEEE} Spoken Language Technology Workshop, {SLT} 2010,
  Berkeley, California, USA, December 12-15, 2010}, pages 19--24. {IEEE}.

\bibitem[{Vaswani et~al.(2017)Vaswani, Shazeer, Parmar, Uszkoreit, Jones,
  Gomez, Kaiser, and Polosukhin}]{vaswani2017attention}
Ashish Vaswani, Noam Shazeer, Niki Parmar, Jakob Uszkoreit, Llion Jones,
  Aidan~N. Gomez, Lukasz Kaiser, and Illia Polosukhin. 2017.
\newblock \href
  {https://proceedings.neurips.cc/paper/2017/hash/3f5ee243547dee91fbd053c1c4a845aa-Abstract.html}
  {Attention is all you need}.
\newblock In \emph{Advances in Neural Information Processing Systems 30: Annual
  Conference on Neural Information Processing Systems 2017, December 4-9, 2017,
  Long Beach, CA, {USA}}, pages 5998--6008.

\bibitem[{Weischedel and Consortium(2013)}]{trove.nla.gov.au/work/192067053}
Ralph~M Weischedel and Linguistic~Data Consortium. 2013.
\newblock Ontonotes release 5.0.
\newblock Title from disc label.

\bibitem[{Williams and Zipser(1989)}]{williams}
Ronald~J. Williams and David Zipser. 1989.
\newblock \href {https://doi.org/10.1162/neco.1989.1.2.270} {A learning
  algorithm for continually running fully recurrent neural networks}.
\newblock \emph{Neural Comput.}, 1(2):270--280.

\bibitem[{Yamada et~al.(2020)Yamada, Asai, Shindo, Takeda, and
  Matsumoto}]{yamada2020luke}
Ikuya Yamada, Akari Asai, Hiroyuki Shindo, Hideaki Takeda, and Yuji Matsumoto.
  2020.
\newblock \href {https://doi.org/10.18653/v1/2020.emnlp-main.523} {{LUKE:} deep
  contextualized entity representations with entity-aware self-attention}.
\newblock In \emph{Proceedings of the 2020 Conference on Empirical Methods in
  Natural Language Processing, {EMNLP} 2020, Online, November 16-20, 2020},
  pages 6442--6454. Association for Computational Linguistics.

\bibitem[{Yan et~al.(2019)Yan, Deng, Li, and Qiu}]{yan2019tener}
Hang Yan, Bocao Deng, Xiaonan Li, and Xipeng Qiu. 2019.
\newblock \href {http://arxiv.org/abs/1911.04474} {{TENER:} adapting
  transformer encoder for named entity recognition}.
\newblock \emph{CoRR}, abs/1911.04474.

\bibitem[{Yang et~al.(2017{\natexlab{a}})Yang, Zhang, and
  Dong}]{yang2017neural}
Jie Yang, Yue Zhang, and Fei Dong. 2017{\natexlab{a}}.
\newblock \href {https://doi.org/10.18653/v1/P17-1078} {Neural word
  segmentation with rich pretraining}.
\newblock In \emph{Proceedings of the 55th Annual Meeting of the Association
  for Computational Linguistics, {ACL} 2017, Vancouver, Canada, July 30 -
  August 4, Volume 1: Long Papers}, pages 839--849. Association for
  Computational Linguistics.

\bibitem[{Yang et~al.(2017{\natexlab{b}})Yang, Salakhutdinov, and
  Cohen}]{yang2017transfer}
Zhilin Yang, Ruslan Salakhutdinov, and William~W. Cohen. 2017{\natexlab{b}}.
\newblock \href {https://openreview.net/forum?id=ByxpMd9lx} {Transfer learning
  for sequence tagging with hierarchical recurrent networks}.
\newblock In \emph{5th International Conference on Learning Representations,
  {ICLR} 2017, Toulon, France, April 24-26, 2017, Conference Track
  Proceedings}. OpenReview.net.

\bibitem[{Zhang et~al.(2018)Zhang, Liu, and Fu}]{zhang2018neural}
Qi~Zhang, Xiaoyu Liu, and Jinlan Fu. 2018.
\newblock \href
  {https://www.aaai.org/ocs/index.php/AAAI/AAAI18/paper/view/16368} {Neural
  networks incorporating dictionaries for chinese word segmentation}.
\newblock In \emph{Proceedings of the Thirty-Second {AAAI} Conference on
  Artificial Intelligence, (AAAI-18), the 30th innovative Applications of
  Artificial Intelligence (IAAI-18), and the 8th {AAAI} Symposium on
  Educational Advances in Artificial Intelligence (EAAI-18), New Orleans,
  Louisiana, USA, February 2-7, 2018}, pages 5682--5689. {AAAI} Press.

\bibitem[{Zhang and Yang(2018)}]{zhang2018chinese}
Yue Zhang and Jie Yang. 2018.
\newblock \href {https://doi.org/10.18653/v1/P18-1144} {Chinese {NER} using
  lattice {LSTM}}.
\newblock In \emph{Proceedings of the 56th Annual Meeting of the Association
  for Computational Linguistics, {ACL} 2018, Melbourne, Australia, July 15-20,
  2018, Volume 1: Long Papers}, pages 1554--1564. Association for Computational
  Linguistics.

\end{thebibliography}
	\bibliographystyle{acl_natbib}
	
	\clearpage
	\appendix
	\renewcommand\thetable{\Alph{section}\arabic{table}}
	
	\onecolumn
	\appendix

	\begin{CJK*}{UTF8}{gbsn}
	\section{Case Study}
	\setcounter{table}{0}
	\begin{table}[h!]
		\renewcommand\arraystretch{0.7}

		\rule[-2pt]{\textwidth}{0.05em}
		
		\begin{tabular}{cccccccc}
			Input(1) & Play & \cclq{this}  & \cclq{is}    & \cclq{colour} & by & panda  & Bear.  \\
			In dict  & o    & \cclw{track} & \cclw{track} & \cclw{track}  & o  & artist & artist \\
			Baseline & o    & \ccle{album} & \ccle{album} & \ccle{album}  & o  & artist & artist \\
			DyLex    & o    & \cclr{track} & \cclr{track} & \cclr{track}  & o  & artist & artist \\
		\end{tabular} 
	
		\rule[-2pt]{\textwidth}{0.05em}
		\begin{tabular}{cccccccccccc}
			Input(2) & Use & netflix & to & play & \cclq{bizzy}  & \cclq{bone}   & kiss  & me    & good-night & Serge-ant & major \\
			In dict  & o   & service & o  & o    & \cclw{artist} & \cclw{artist} & track & track & track & track & track \\
			Baseline & o   & service & o  & o    & \ccle{track}  & \ccle{track}  & track & track & track & track & track \\
			DyLex    & o   & service & o  & o    & \cclr{artist} & \cclr{artist} & track & track & track & track & track \\
		\end{tabular}
		
		\rule[-2pt]{\textwidth}{0.05em}
		\scalebox{0.90}{
			\begin{tabular}{cccccccccccccccc}
				Input(3) & I & want & to & add & \cclq{hind}   & \cclq{etin}   & to & my & la  & mejor& musica   & dance    & 2017  & playlist \\
				In dict  & o & o    & o  & o   & \cclw{entity} & \cclw{entity} & o  & owner & plst & plst & plst & plst & plst & plst \\
				Baseline & o & o    & o  & o   & \ccle{artist} & \ccle{artist} & o  & owner & plst & plst & plst & plst & plst & plst \\
				DyLex    & o & o    & o  & o   & \cclr{artist} & \cclr{artist} & o  & owner & plst & plst & plst & plst & plst & plst
		\end{tabular}}
		\rule[-2pt]{\textwidth}{0.05em}
		\begin{tabular}{cccccccccccc}
			Input(4) & what & is & the & weather & like & in & \cclq{north}   & \cclq{salt}    & \cclq{lake}    & and & afghanistan \\
			In dict  & o    & o  & o   & o       & o    & o  & \cclw{city}    & \cclw{city}    & \cclw{city}    & o   & country     \\
			Baseline & o    & o  & o   & o       & o    & o  & \ccle{country} & \ccle{country} & \ccle{country} & o   & country     \\
			DyLex    & o    & o  & o   & o       & o    & o  & \cclr{city}    & \cclr{city}    & \cclr{city}    & o   & country    
		\end{tabular}
		
		\rule[-2pt]{\textwidth}{0.05em}
		\scalebox{1.0}{
			\begin{tabular}{ccccccccccc}
				Input(5) & I & want & to & book & a & cafe      & for & 3 & in & \cclq{fargo}   \\
				In dict  & o & o    & o  & o    & o & res\_type & o   & o & o  & \cclw{city}    \\
				Baseline & o & o    & o  & o    & o & res\_type & o   & o & o  & \ccle{country} \\
				DyLex    & o & o    & o  & o    & o & res\_type & o   & o & o  & \cclr{city}   
			\end{tabular}
		}
		
		\rule[-2pt]{\textwidth}{0.05em}
		\begin{tabular}{cccccccc}
			Input(6) & play & \cclq{the} 	& \cclq{new} 	& \cclq{noise} 	& \cclq{theology} 	& \cclq{ep} 	& \\
			In dict  & o  	& \cclw{object}	& \cclw{object} & \cclw{object} & \cclw{object} 	& \cclw{object} & \multirow{-2}{*}{intent} \\
			Baseline & o 	& \ccle{plst} 	& \ccle{plst} 	& \ccle{plst} 	& \ccle{plst} 		& \ccle{plst} 	& PlayMusic                \\
			DyLex    & o 	& \cclr{object} & \cclr{object} & \cclr{object} & \cclr{object} 	& \cclr{object} & SearchCreativeWork      
		\end{tabular}
		
		\rule[-2pt]{\textwidth}{0.05em}
		\begin{tabular}{cccccccccc}
			Input(7) & Find & \cclq{a}      & \cclq{man}    & \cclq{needs}  & \cclq{a}      & \cclq{maid}   & \cclq{Bear.}  & \multirow{4}{*}{} & \multirow{2}{*}{Intent} \\
			In dict  & o    & \cclw{object} & \cclw{object} & \cclw{object} & \cclw{object} & \cclw{object} & \cclw{object} & &   \\
			Baseline & o    & \ccle{movie}  & \ccle{movie}  & o  			& \ccle{movie}  & \ccle{movie}  & \ccle{movie}&& SearchScreeningEvent  \\
			DyLex    & o    & \cclr{object} & \cclr{object} & \cclr{object} & \cclr{object} & \cclr{object} & \cclr{object} & & SearchCreativeWork   
		\end{tabular}
		
		\rule[-2pt]{\textwidth}{0.05em}
		\scalebox{1.0}{
			\begin{tabular}{ccccccccccc}
				Input(8) & 播 	& 放 & \cclq{林}		& \cclq{星} & \cclq{辰} & 的  & \cclq{音} & \cclq{乐} & \cclq{盒} \\
				In dict  & o  	& o  & \cclw{artist}  & \cclw{artist}  & \cclw{artist}  & o & \cclw{track} & \cclw{track} & \cclw{track}  & \multirow{-2}{*}{Intent} \\
				Baseline & o    & o  & \ccle{artist} & \ccle{artist} & \ccle{track} & \ccle{track}   & \ccle{track}& \ccle{track} &   \ccle{track}   & PlayMusic       \\
				DyLex    & o    & o  & \cclr{artist}  & \cclr{artist}  & \cclr{artist}  & o & \cclr{track} & \cclr{track} &  \cclr{track} & PlayMusic    
			\end{tabular}
		}
		
		\rule[-2pt]{\textwidth}{0.05em}
		\scalebox{1.0}{
			\begin{tabular}{ccccccccccc}
				Input(9) & 播 & 放 & \cclq{林}& \cclq{星} & \cclq{辰} & \cclq{的}  & \cclq{音} & \cclq{乐} & \cclq{盒} \\
				In dict  & o    & o  & \cclw{track}  & \cclw{track}  & \cclw{track}  & \cclw{track} & \cclw{track} & \cclw{track} & \cclw{track}  & \multirow{-2}{*}{Intent} \\
				Baseline & o    & o  & \ccle{artist} & \ccle{artist} & \ccle{track} & \ccle{track}   & \ccle{track}& \ccle{track} &   \ccle{track}   & PlayMusic       \\
				DyLex    & o    & o  & \cclr{track}  & \cclr{track}  & \cclr{track}  & \cclr{track} & \cclr{track} & \cclr{track} &  \cclr{track} & PlayMusic    
			\end{tabular}
		}
		
		\rule[-2pt]{\textwidth}{0.05em}
		\scalebox{0.97}{
			\begin{tabular}{ccccccccccccccccccc}
				Input(10) & 外 & 国 & 政  & 要 & 发  & 表& 新 & 年& 贺 & 词& 满  & 怀 & 信 & 心  & \cclq{应}& \cclq{对} & 挑 & 战 \\
				In dict  & B & I & B & I & B & I & B & I & B & I & B & I & I & I & \cclw{B} & \cclw{I} & B & I \\
				Baseline & B & I & B & I & B & I & B & I & B & I & B & I & I & I & \ccle{B} & \ccle{B} & B & I \\
				DyLex    & B & I & B & I & B & I & B & I & B & I & B & I & I & I & \cclr{B} & \cclr{I} & B & I \\
		\end{tabular}}
		\rule[-2pt]{\textwidth}{0.05em}
		\scalebox{1.0}{
			\begin{tabular}{cccccccccccccc}
				Input(11) & 环 & \cclq{南} & \cclq{中} &\cclq{国} & \cclq{海} & 自 & 行 & 车 & 赛 & 落 & 幕 & 澳 & 门 \\
				In dict   & B  & \cclw{B} & \cclw{I} & \cclw{I} & \cclw{I} & B & I & I & I & B & I & B & I \\
				Baseline  & B  & \ccle{B} & \ccle{B} & \ccle{I} & \ccle{I} & B & I & I & I & B & I & B & I \\
				DyLex     & B  & \cclr{B} & \cclr{I} & \cclr{I} & \cclr{I} & B & I & I & I & B & I & B & I
			\end{tabular}
		}
		\rule[-2pt]{\textwidth}{0.05em}
		\scalebox{1.0}{
			\begin{tabular}{cccccccccccccccccc}
				Input(12) & 这 & 起 & 发 & 生 & 在 & 校 & 园 & 内 & 的 & 重 & 大 & 安 & 全 & \cclq{责} & \cclq{任} & \cclq{事} & \cclq{故} \\
				In dict   & B & B & B & I & B & B & I & B & B & B & I & B & I & \cclw{B} & \cclw{I} & \cclw{I} & \cclw{I} \\
				Baseline  & B & B & B & I & B & B & I & B & B & B & I & B & I & \ccle{B} & \ccle{I} & \ccle{B} & \ccle{I} \\
				DyLex     & B & B & B & I & B & B & I & B & B & B & I & B & I & \cclr{B} & \cclr{I} & \cclr{I} & \cclr{I}
			\end{tabular}
		}
		\rule[-2pt]{\textwidth}{0.05em}
		\label{tb:case-study}

	\end{table}

	As showed in above, we randomly select some examples of inconsistent predictions before and after adding the lexicons, example [1-5] is from NER, example [6-9] is from NLU, and example [10-12] is from CWS . Each example contains the input sentence, the related matching result, the baseline prediction, and \textbf{DyLex} prediction. Highlighted parts indicate inconsistent results. 
	We make some interesting observations. 
	
	\textbf{CASE \uppercase\expandafter{\romannumeral1}}~~~ Different type of entities can be placed under a same context. For example [1], ``play'' can be followed by TRACK or ALBUM (play [XX]). Model would be confused of whether XX is a TRACK or a ALBUM. In this case lexicons can provide enough type information to acquire a correct result.
	
	\textbf{CASE \uppercase\expandafter{\romannumeral2}}~~~ Chinese word segmentation granularity is flexible according to the context. ``南中国海 (South China Sea)'' can be segmented into ``南 (South)'' and ``中国海 (China Sea)'', or it can be regarded as a single word [11]. At this point, an external lexicon will be benefit for controlling the granularity.
	
	\textbf{CASE \uppercase\expandafter{\romannumeral3}}~~~ It happens that the word combination in slot have different interpretations, usually when the length of a slot is too long. That may cause the discontinuity of slot extraction. For example, we can see an improper O is inserted in the baseline prediction [7]. By incorporating lexicons, the boundary information can enhance the integrity of slot extraction. 
	
	\textbf{CASE \uppercase\expandafter{\romannumeral4}}~~~  Dylex can adapt its prediction to updating lexicons. As example[8-9] illustrated, given different lexicon entries, our framework can understand what ``林星辰的音乐盒'' is, then  dynamicly provide correct slot.
	
	\end{CJK*}

	\section{Lexicon size uesd in different experiment}
	\label{appendix:lexicon_size_exp}
	\setcounter{table}{0}
	\begin{table}[htbp]
		\begin{tabular}{p{3.2cm}p{3cm}p{4cm}<{\centering}p{4cm}<{\centering}}
			\toprule
			Task                    & Datasets       & Exp-Dict         & Sp-Dict      \\
			\midrule
			\multirow{2}{*}{CWS}    & PKU            & 570K             & 57.7K        \\
									& CITYU          & 579K             & 70.5K        \\
			\midrule
			\multirow{4}{*}{NER-CN} & Ontonotes      & 97.2K            & 68.6K        \\
									& MSRA           & 98.1K            & 80.5K        \\
									& Resume         & 97.9K            & 68.9K        \\
									& Weibo          & 96.9K            & 62.9K        \\
			\midrule
			\multirow{2}{*}{NER-EN} & Conll2003      & 1.3M             & 33K          \\
									& Ontonotes5.0   & 1.3M             & 47K          \\
			\midrule
			\multirow{3}{*}{NLU}    & ATIS           & 1.3K             & 1.3K         \\
									& Snips          & 12K              & 12K          \\
									& Industrial NLU & 16M              & 16M          \\
			\bottomrule
		\end{tabular}
		\caption{Lexicon size(number of term) uesd in different experiment}
	\end{table}

\section{Overiew of NER dataset}
	\setcounter{table}{0}
\begin{table}[h]
	\centering
	\scalebox{1.0}{
		\begin{tabular}{lccccccc}
			\toprule
			& Ontones & MSRA & Resume & Weibo & Conll2003 & OntoNotes5.0 \\ 
			\midrule
			train & 15,470 & 46,675 & 3,821 & 1,350 & 14,987 & 115,812 \\ 
			char$_{avg}$ & 36.92 & 45.87 & 32.15 & 54.37 & - & -  \\ \
			word$_{avg}$ & 17.59 & 22.38 & 24.99 & 21.49 & 13.5 & 9.40 \\ 
			entity$_{avg}$ & 1.15 & 1.58 & 3.48 & 1.42 & 1.56 & 0.71  \\ 
			\bottomrule
		\end{tabular}
	}
	\caption{Overiew of NER dataset}
	\label{tb:overview-ner-dataset}
\end{table}
	
\section{Overiew of NLU dataset}
	\label{appedix:nlu-dataset}
	\setcounter{table}{0}
	\begin{table}[htbp]
	\centering
	\scalebox{1.0}{
		\begin{tabular}{lcccccc}
			\toprule
			Type & Dataset & Train & Dev & Test & Intents & Slots \\ 
			\midrule
			Industrial & - & 80,000 & 30,000 & 30,000 & 500 & 400 \\ 
			\midrule
			\multirow{2}{*}{Public} & 
			Snips & 13,084 & 700 & 700 & 7 & 72  \\
			& ATIS & 4,478 & 500 & 893 & 21 & 120  \\ 
			\bottomrule
		\end{tabular}
	}
	\mycaption{The stastics of NLU datasets.}
	\label{tb:NLU-dataset}
\end{table}
	The Chinese industrial NLU dataset is a corpus specially used to train mobile phone assistants. The data set includes 80k Training set, 30k Dev set and 30k Test set. The annotation contains 500 types of intentions commonly used by mobile assistants , which are divided into 8 categories such as setting and control. There are 400 slots categories in total. The data is labeled using crowdsourcing. The cost is about 1
    dollar per sentence. Each sentence was marked by 3 people, and finally the result was determined by voting. At last, there is a acceptance sampling, and professionals will spot check the quality of each batch, and the error is controlled within 1\%.
	
\section{A concrete example of a lexicon}
    \label{app:fragment of lexicon}
    \setcounter{table}{0}
	\begin{table}[h]
	    \centering
	    \setlength{\tabcolsep}{10mm}{
        \begin{tabular}{lc}
            \toprule
            Item & Category \\
            \midrule
            cathy mu $\sim$no \#\#z                & PER  \\ \hline
            pieter pieter \#\#sz barbie \#\#rs     & PER  \\ \hline
            bell high school                       & ORG  \\ \hline
            fredrik ri \#\#sp                      & PER  \\ \hline
            liverpool                              & ORG  \\ \hline
            venice gardens                         & LOC  \\ \hline
            brant \#\#ford golden eagles           & ORG  \\ \hline
            jerry and \#\#rus                      & PER  \\ \hline
            taylor leon                            & PER  \\ \hline
            kata \#\#rina e \#\#wer \#\#lo \#\#f   & PER  \\ \hline
            anne finch                             & PER  \\ \hline
            hanna st \#\#yre \#\#ll                & PER  \\ \hline
            the big blue                           & MISC \\ \hline
            math \#\#are united                    & ORG  \\ \hline
            var \#\#ana \#\#si college of pharmacy & ORG  \\ \hline
            gilbert \#\#sville                     & LOC  \\ 
            \bottomrule
        \end{tabular}
        }
    \caption{
        A fragment of the lexicon used in this article. The Item on the left is the wordpiece of the words, and the corresponding category on the right.
    }
\end{table}

\section{Hyperparameters}
\label{app:hyper-param}

\begin{table}[H]
    \centering
    \setcounter{table}{0}
    \begin{tabular}{lll}
        \toprule
        batch\_size        & [32, 64]    &   \\
        learning\_rate     & 2e-5  &         \\
        optimizer          & Adam &          \\
        weight\_decay      & 0.01  &         \\
        dropout            & 0.1    &         \\
        max\_seq\_length   & 128   &         \\
        dict\_candidate    & 16    & \#the maximum number of matches per sentence  \\
        top\_n             & 1     & \#number of matches reserved for each position   \\
        warmup\_proportion & 0.1   &     \\
        epochs             & 20    &      \\
        use\_first         & True  & \#only the first character category is used to predict   the entity type \\
        \bottomrule
\end{tabular}
\caption{
    The hyperparameters used in the experiment. Other hyperparameters default are consistent with BERT.
}
\end{table}

\end{document}